\pdfoutput=1

\documentclass[11pt]{article}

\usepackage{emnlp2021}

\usepackage{times}
\usepackage{latexsym}

\usepackage[T1]{fontenc}

\usepackage[utf8]{inputenc}
\usepackage{graphicx}
\usepackage{multirow}
\usepackage{tabularx}
\usepackage{caption}
\usepackage{subcaption}

\usepackage{microtype}
\usepackage{makecell}
\usepackage{amssymb}
\usepackage{pifont}
\newcommand{\cmark}{\ding{51}}%
\newcommand{\xmark}{\ding{55}}%
\renewcommand{\UrlFont}{\ttfamily}
\usepackage[colorinlistoftodos]{todonotes}
\newcounter{mycomment}

\makeatletter
\def\thanks#1{\protected@xdef\@thanks{\@thanks
        \protect\footnotetext{* #1}}}
\makeatother

\title{MultiDoc2Dial: Modeling Dialogues Grounded in Multiple Documents}

\author{Song Feng\thanks{Equal contribution.}\footnotemark[1] \\ IBM Research AI \\ \texttt{sfeng@us.ibm.com} \And
Siva Sankalp Patel\footnotemark[1] \\ IBM Research AI \\ \texttt{siva.sankalp.patel@ibm.com} \AND
Hui Wan \\ IBM Research AI \\ \texttt{hwan@us.ibm.com} \And
Sachindra Joshi \\ IBM Research AI \\ \texttt{jsachind@in.ibm.com}}

\begin{document}
\maketitle
\begin{abstract}
We propose MultiDoc2Dial, a new task and dataset on modeling goal-oriented dialogues grounded in multiple documents. Most previous works treat document-grounded dialogue modeling as a machine reading comprehension task based on a single given document or passage. In this work, we aim to address more realistic scenarios where a goal-oriented information-seeking conversation involves multiple topics, and hence is grounded on different documents. To facilitate such a task, we introduce a new dataset that contains dialogues grounded in multiple documents from four different domains. We also explore modeling the dialogue-based and document-based context in the dataset. We present strong baseline approaches and various experimental results, aiming to support further research efforts on such a task.

\end{abstract}

\section{Introduction}
With the recent advancements in NLP, there has been a surge of research interests and efforts in developing conversational systems for various domains. An important task in the field is conversational question answering and document-grounded dialogue modeling. Prior work typically formulates the task as a machine reading comprehension task assuming the associated document or text snippet is given, such as QuAC~\citep{choi-etal-2018-quac}, ShARC~\citep{saeidi-etal-2018-interpretation}, CoQA~\citep{reddy-etal-2019-coqa}, OR-QuAC~\citep{10.1145/3397271.3401110} and Doc2Dial \citep{feng-etal-2020-doc2dial}. However, such task setup neglects the common real-life scenarios where a goal-oriented conversation could correspond to several sub-goals that are addressed in different documents. 
In this work, we propose a new task and dataset, \textbf{MultiDoc2Dial}, on modeling goal-oriented dialogues that are grounded in multiple documents. 

\begin{figure*}[!t]
  \centering
  \includegraphics[width=1.0\textwidth]{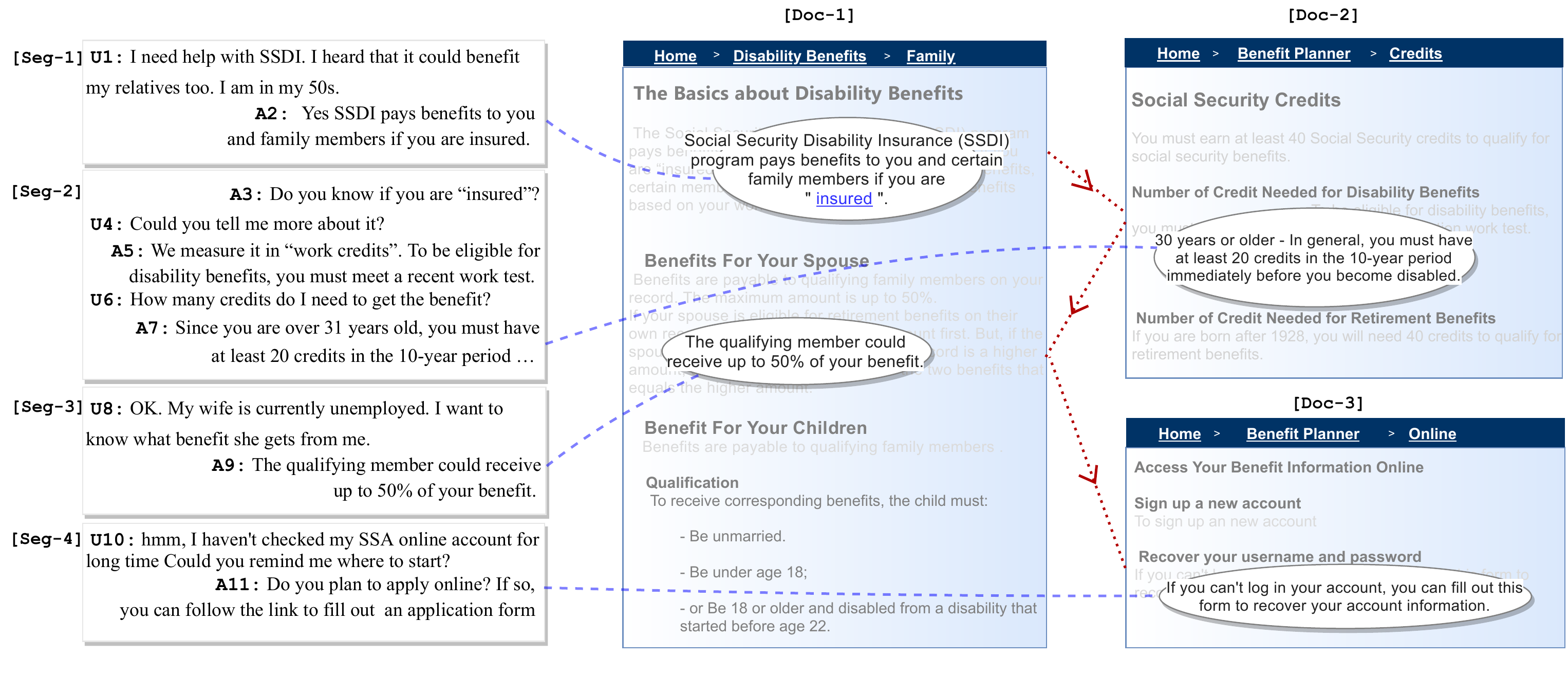}
  \caption{\label{fig:intro} A sample goal-oriented dialogue (on the left) that is grounded in three relevant documents (on the right).} 
\end{figure*}

We illustrate the proposed task in Figure \ref{fig:intro}. It includes a goal-oriented dialogue with four segments on the left and three relevant documents on the right. Each dialogue segment indicates that all turns within it are grounded in a same document, e.g., turns from A3 to A7 in \texttt{Seg-2} are all grounded in \texttt{Doc-2}. The blue dashed lines connect a dialogue turn with its corresponding relevant span in a document. The red dotted lines with arrows indicate that the dialogue flow shifts among the grounding documents through the conversation, i.e., \texttt{Doc-1} $\to$ \texttt{Doc-2} $\to$ \texttt{Doc-1} $\to$ \texttt{Doc-3}. This example highlights certain challenges for identifying the relevant grounding content among different documents dynamically in a conversation. For instance, agent response A2 mentions `insured' as an important condition based on a span in \texttt{Doc-1}. However, there are no more details about `insured' in \texttt{Doc-1}. To further discuss about `insured`, the conversation naturally switches to refer to another document \texttt{Doc-2}. Another challenge is to handle deep dialogue context. For instance, to provide response to U4 or U6 in \texttt{Seg-2}, the agent needs to understand the context of `disability benefit' mentioned in \texttt{Seg-1}. There are also cases such as \texttt{Seg-4} where its turn U10 starts a new question related to \texttt{Doc-3} but seems independent of the previous segments. The task goal in the example is naturally simple, but it still reveals the realistic expectation of document-grounded dialogue modeling that is still yet to be met.

To the best of our knowledge, there is no existing task or dataset that addresses the scenarios where the grounding documents of goal-oriented dialogues are unknown and dynamic. To facilitate the study in this direction, we introduce a new dataset that contains conversations that are grounded in multiple documents. To construct the dialogue flows that involve multiple documents, we derive a new approach from the data collection pipeline proposed in \citet{Feng_Fadnis_Liao_Lastras_2020}. The newly composed dialogue flows have multiple dialogue segments where two adjacent segments are grounded in different documents. We harvest original utterances whenever possible and collect new utterances when necessary. The dataset includes about 4800 dialogues with an average of 14 turns that are grounded in 488 documents from four different domains. 

Inspired by recent advances in open retrieval question answering (QA) tasks \citep{lee-etal-2019-latent,guu2020realm,karpukhin-etal-2020-dense,lewis2020retrieval,khattab2020relevance}, we develop baseline models based on the retriever-reader architecture \citep{karpukhin-etal-2020-dense,lewis2020retrieval}. Compared to the existing open retrieval QA \citep{lee-etal-2019-latent,min-etal-2020-ambigqa} and open retrieval conversational QA tasks \citep{qu2020open,li2021graph}, our dataset contains more complex and diverse dialogue scenarios based on diversified documents from multiple domains. 
To work towards modeling the interconnected contexts from dialogues and documents, we utilize document and dialog-based structure information. For the former, we segment a document into passages while maintaining its hierarchical contextual information. For the latter, in addition to combining current turn and dialogue history~\citep{qu2020open,qu2021weaklysupervised}, we also experiment with different ways to encode the current turn separately based on the intuition that the latest turn, with a change in topic, could be semantically distant from the dialogue history.
We also explore different retriever settings in our experiments. 

We propose two tasks for modeling dialogues that are grounded in multiple documents. One is to generate the grounding document span; and the other is to generate agent response given current turn, dialogue history and a set of documents. We evaluate the performances of the retriever and generator in baseline models trained on MultiDoc2Dial dataset.

\begin{table}
\begin{center}
\begin{tabular}
{ p{2.2cm} c c c c c}
\hline
\textbf{Data\&Task} & \textbf{OB} & \textbf{GO} & \textbf{GA} & \textbf{FD} & \textbf{MD}  \\ \hline
ShARC & \xmark & \cmark  & \xmark & \xmark & \xmark   \\
DoQA & \xmark & \cmark   & \xmark & \cmark & \xmark   \\
Doc2Dial & \xmark & \cmark & \cmark & \cmark & \xmark \\
QuAC  & \xmark & \xmark   & \xmark & \cmark & \xmark \\
OR-QuAC & \cmark & \xmark & \xmark & \cmark & \xmark \\
MultiDoc2Dial & \cmark & \cmark & \cmark & \cmark & \cmark \\
\hline
\end{tabular}
\caption{Comparison among selected datasets and tasks on document-grounded dialogue and conversational QA on different aspects: open-book (OB), goal-oriented (GO), grounding available (GA), full-document (FD), multiple-document (MD).
}
\label{tab:comparison}
\end{center}
\end{table}

We summarize our contributions as follows:
\begin{itemize}
    \item We propose a novel task and dataset, called \textbf{MultiDoc2Dial}, on modeling goal-oriented dialogues that are grounded in multiple documents from various domains. We aim to challenge recent advances in dialogue modeling with more realistic scenarios that is hardly addressed in prior work.
    
    \item We present strong baseline approaches and evaluations for two tasks based on MultiDoc2Dial dataset in hope to support future significant research effort in the direction. Our data and code are available at \url{https://doc2dial.github.io/multidoc2dial/}. 
\end{itemize}

\section{Multi-Document-Grounded Dialogue}
\subsection{Data}
We present MultiDoc2Dial, a new dataset that contains 4796 conversations with an average of 14 turns grounded in 488 documents from four domains. This dataset is constructed based on Doc2Dial dataset V1.0.1 \footnote{\url{https://doc2dial.github.io/data.html}}. MultiDoc2Dial shares the same set of annotations as Doc2Dial. For document data, it includes HTML mark-ups such as list, title and document section information as shown in Figure \ref{fig:intro}. For dialogue data, each dialogue turn is annotated with role, dialogue act, human-generated utterance and the grounding span with document information. Each dialogue contains one or multiple segment where each indicates that all turns within one segment are grounded in a same document. For instance, the dialogue in Figure \ref{fig:intro} has four segments that are grounded in three documents. 

We exclude the `irrelevant' scenarios in Doc2Dial where the user question is unanswerable and leave it for future work. We also filter out certain dialogues when we identify more than four noisy turns per dialogue. There is a total of 61078 dialogue turns in MultiDoc2Dial dataset, which consists of 38\% user questions, 12\% agent follow-up questions and the rest as responding turns. Table \ref{tab:data} shows the statistics of the dataset by domain, including the number of dialogues with two segments (two-seg), more than two segments ({>two-seg}), and no segmentations (single). 


To create the data, we derive a new data construction approach from the pipelined framework by \citet{Feng_Fadnis_Liao_Lastras_2020}. We first create dialogue flows that correspond to multiple documents and then re-collect the utterances for certain turns based on dialogue scenes in the given flow via crowdsourcing. In addition, we aim to reuse previous turns from doc2dial dataset wherever possible and collect new turns when necessary to compose the new dialogues. 

\subsubsection{Dialogue Flow}
\label{sec:data_flow}
To construct dialogue flows grounded in multiple documents, we need to split the existing dialogues into segments and recompose them. The main idea is to identify the position where the previous topic can possibly end and then find a segment with a new topic that is grounded in a different document, for which we utilize both document-based and dialogue-based structure knowledge. 

\paragraph{Dialogue Segmentation} To segment dialogues, we identify all the candidate splitting positions
based on dialogue act and turn index. Intuitively, we aim to maintain semantic coherence of a dialogue segment \citep{mele2020topic}. Thus, we only split after an agent turn with dialogue act as `responding with an answer' while the next turn is not `asking a follow-up question'. We randomly select a number of splitting positions per existing dialogue and obtain 2 to 4 segments per dialogue.

\paragraph{Document Transition} To simulate the document-level topic shift between dialogue segments, we identify different types of grounding document transition in dialogue, including (1) the following grounding document is explicitly closely related to the preceding grounding document, such as \texttt{Seg-1} and \texttt{Seg-2} in Figure~\ref{fig:intro}; and (2) the two documents are not necessarily closely related, such as \texttt{Seg-3} and \texttt{Seg-4}. For the former case, we exploit document-based structure knowledge of a domain to determine the semantic proximity of document pairs, including (1) document-level hierarchical structure indicated by the website URLs, such as \texttt{Doc-2} and \texttt{Doc-3} shares one parent topic; and (2) hyperlinks between pages, such as, hyperlink of `insured' in \texttt{Doc-1} to \texttt{Doc-2}. For the latter case, we just randomly select document pairs from the same domain if they do not belong to the former case.

\paragraph{Re-composition} Last, we combine multiple dialogue segments to form a new dialogue flow based on the following rules: (1) a dialogue segment can only appear in one new dialogue flow; (2) the grounding documents of two adjacent dialogue segments must be different; (3) we keep the new dialogue flows between 6 and 20 turns by filtering out shorter dialogues or discarding later turns of longer dialogues.

\begin{table}
\begin{center}
\begin{tabular}
{ m{1cm} m{0.8cm} m{0.8cm} m{0.8cm}  m{0.8cm} m{0.8cm}  }
\hline
\textbf{domain} & \textbf{\#doc} &  \textbf{\#dial} & \textbf{two-seg} & \textbf{>two-seg} & \textbf{single}\\
\hline
ssa & 109 & 1191 & 701 & 188 & 302 \\
va  & 138 & 1337 & 648 & 491 & 198 \\
dmv & 149 & 1328 & 781 & 257 & 290 \\
student & 92 & 940 & 508 & 274 & 158  \\ \hline
total & 488 & 4796 & 2638 & 1210 & 948 \\
\hline

\end{tabular}
\caption{MultiDoc2Dial data statistics.}
\label{tab:data}
\end{center}
\end{table}

\subsubsection{Data Collection}
After we re-compose dialogue flows with multiple segments, we need to re-write certain dialogue turns since some of the original turns could be under-specified when taken out of the previous context, especially when they are re-positioned at the beginning of a dialogue segment. For instance, if we use \texttt{Seg-3} in a new dialogue context, then we expect \texttt{U8} to be enhanced with necessary background such as ``I am qualified for disability benefit. My wife is currently unemployed. I want to know what benefit she gets from me'' based on \texttt{Doc-1}. 

To collect the rewriting of a given dialogue turn, we provide context information including up to four preceding turns, the succeeding turn, the associated document title information and the grounding span in a document section. We ask crowdsourced contributors to rewrite an utterance fit for the given context, by adding necessary background information and removing the irrelevant or contradicting content. For quality control, we also insert various template-based placeholders for the crowd to modify accordingly. The task would be rejected if they fail to modify the placeholders. We collect over 6000 turns for improving the multi-segmented dialogues. The task was performed by 30 qualified contributors from {\UrlFont{appen.com}}. More information about the crowdsourcing task can be found in Appendix \ref{appx_crowd}. 


\begin{table}
\begin{center}
\begin{tabular}
{ l c c c }
\hline
\textbf{} & \textbf{Train} & \textbf{Val} &  \textbf{Test}  \\ \hline
\#dialogues & 3474 & 661 & 661 \\
\#queries & 21453 & 4201 & 4094 \\ 
avg query length &  104.6 & 104.2 & 96.5 \\
avg response length &  22.8 & 21.6 & 22.3 \\
\hline
\#passages (token) & \multicolumn{3}{c}{4283}\\
\#passages (struct) & \multicolumn{3}{c}{4110}\\ 
avg length (token) & \multicolumn{3}{c}{100} \\
avg length (struct) & \multicolumn{3}{c}{106.6} \\
\hline
\end{tabular}
\caption{Data statistics of document passages and dialogue data based on splits. The average length is based on the number of tokens.
}
\label{tab:split}
\end{center}
\end{table}

\subsection{Tasks}
We propose two tasks for the evaluations on MultiDoc2Dial dataset.
\subsubsection{Task I: Grounding Span Prediction}
This task aims to predict the grounding document span for the next agent response. The input includes (1) current user turn, (2) dialogue history, and (3) the entire set of documents from all domains. The target output is a grounding text span from one document that is relevant to the next agent response. To train a dialogue system to be able to provide Fine-grained grounding information can be an important step for improving the interpretability and trustworthiness of neural-model-based conversational systems \citep{carvalho2019machine}.

\subsubsection{Task II: Agent Response Generation}
In this task, we aim to generate the next agent response, which includes asking follow-up questions or providing an answer to a user question. Again, the input includes (1) current user turn, (2) dialogue history and (3) the entire set of documents from all domains. The target output is the next agent response in natural language. This task is considered a more difficult task than Task I since agent utterance varies in style and not directly extracted from document content. A related task is to simulate user utterances, which could be a more challenging task, since they are even more diversified in style and content. We leave this task for future work.

\section{Model}
We propose to formulate the tasks on predicting agent turn as an end-to-end generation task inspired by the recent development of retriever-reader architecture \citep{karpukhin-etal-2020-dense,lewis2020retrieval}. We consider Retrieval-Augmented Generation (RAG) \citep{lewis2020retrieval} as the base model. It includes a retriever component and a generative reader component. The retriever aims to retrieve the most relevant document passages given dialogue query; the generator, a pre-trained seq2seq model in our case, takes the combined dialogue query and top-n document passages as input and generates the target output.

Since we need to deal with the contextual information of dialogue turns and document passages, we further investigate how to utilize dialogue-based and document-based structure information for the retriever component. 

\subsection{Document-based Structure}
\label{model-doc}
Previous approaches for open retrieval question answering typically split long documents into smaller text passages by sliding window. In this work, we investigate two ways of segmenting document content: (1) we also split a document based on a sliding window size of N tokens; (2) we utilize document structural information indicated by markup tags in HTML files. 
Inspired by document tree structures used in \citet{Feng_Fadnis_Liao_Lastras_2020, wan-etal-2021-structure}, we segment the document based on original paragraphs indicated by mark-up tags such as \texttt{<p>} or \texttt{<ul>} and then attach the hierarchical titles to each paragraph as a passage, e.g., adding `The Basics about Disability Benefits / Benefits for Your Children / Qualification` to the last paragraph in \texttt{[Doc-1]} in Figure \ref{fig:intro}. 

\subsection{Dialogue-based Structure}
\label{model-dpr}
For dialogues with multiple topic-based segments, a turn can be more distant from previous turns lexically and semantically when the topic shifts at the turn \cite{arguello-rose-2006-topic}. Thus, we also consider incorporating the retrieval results only by the current turn in addition to the retrieval results by the combination of current turn and history \citep{qu2020open,qu2021weaklysupervised}. To obtain the representation for current turn, we experiment with two ways for BERT-based question encoder in RAG model. One is based on the common \texttt{[CLS]} token embeddings; one is based on pooled token embeddings \citep{choi2021evaluation}.

\section{Experiments}

\begin{table*}[ht!]
\centering
\begin{tabular}{l c c c  c c c |c c c  c c c }
\hline
& \textbf{F1} & \textbf{EM} & \textbf{BL} & \textbf{@1} & \textbf{@5}&  \textbf{@10}
& \textbf{F1} & \textbf{EM} & \textbf{BL} & \textbf{@1} & \textbf{@5}&  \textbf{@10}
\\ \hline
& \multicolumn{6}{c|}{test set} &  \multicolumn{6}{c}{validation set} \\
\hline
D\textsuperscript{token}-bm25 & 36.3 & 19.0 & 24.2 & 20.5 & 41.6 & 50.3 & 37.3 & 19.5 & 25.4 & 19.6 & 41.0 & 48.5 \\
D\textsuperscript{struct}-bm25 & 30.5 & 14.5 & 21.4 & 18.0 & 42.5 & 53.0 & 31.1 & 15.5 & 22.1 & 19.6 & 42.0 & 50.8 \\
\hline
D\textsuperscript{token}-nq & 40.0 & 22.3 & 28.1 & 27.7 & 54.3 & 64.5 & 38.7 & 20.8 & 27.0 & 27.9 & 51.3 & 60.7 \\
D\textsuperscript{struct}-nq & 39.8 & 22.3 & 28.7 & 28.6 & 54.0 & 64.2 & 39.4 & 22.1 & 27.7 & 28.5 & 52.9 & 62.1 \\
\hline
D\textsuperscript{token}-ft & 43.6 & 26.4 & 35.1 & 36.4 & 68.1 & 77.9 & 41.2 & 24.7 & 33.3 & 35.3 & 63.7 & 73.5 \\
D\textsuperscript{struct}-ft & 43.5 & 26.1 & 34.5 & 39.1 & \textbf{69.4} & \textbf{78.9} & 41.8 & 24.8 & 32.4 & 38.0 & \textbf{66.2} & \textbf{75.3} \\
D\textsuperscript{token}-rr-cls-ft & 42.1 & 25.0 & 33.5 & 35.9 & 67.0 & 76.9 & 40.5 & 24.1 & 32.4 & 35.4 & 62.9 & 72.4 \\
D\textsuperscript{struct}-rr-cls-ft & 43.5 & 26.2 & 34.0 & 37.3 & 67.9 & 78.0 & 41.7 & \textbf{24.9} & 31.5 & 37.2 & 64.7 & 74.1 \\
D\textsuperscript{token}-rr-pl-ft & 43.5 & \textbf{26.6} & \textbf{35.4} & 36.1 & 67.3 & 77.6 & 41.7 & \textbf{24.9} & \textbf{33.7} & 34.8 & 62.7 & 72.3 \\
D\textsuperscript{struct}-rr-pl-ft & \textbf{43.7} & 26.3 & 34.4 & \textbf{39.3} & 69.1 & 78.7 & \textbf{41.9} & \textbf{24.9} & 32.4 & \textbf{38.4} & 65.9 & 75.2 \\
\hline

\end{tabular}
\caption{Evaluation results of Task I on grounding span generation task.}
\label{tab:task1}
\end{table*}

\begin{table}
\begin{center}
\begin{tabular}
{ c c c c c}
\hline
\textbf{} & \textbf{Doc-Seg} & \textbf{@1} &  \textbf{@5} & \textbf{@10} \\
\hline
\multirow{2}{3.5em}{BM25} & token & 19.5 & 42.7 & 51.4\\
  & struct & 19.6 & 41.9 & 50.8 \\
\hline
\multirow{2}{3.5em}{DPR-nq}  & token & 13.5 & 30.2 & 38.9\\
& struct  & 15.5 & 35.3  & 45.6 \\
\hline
\multirow{2}{3.5em}{DPR-ft}  &  token & 36.5 & 57.4 & 64.6 \\
& struct  & \textbf{49.0} & \textbf{72.3} &  \textbf{80.0}\\
\hline

\end{tabular}
\caption{Retrieval results on validation set.}
\label{tab:ir}
\end{center}
\end{table}

We evaluate the proposed approaches on two tasks for predicting next agent turn, i.e., grounding generation and response generation. Given current turn, dialogue history and all available documents in the dataset, we aim to evaluate generated text along with intermediate retrieval results. 
We split the data into train/validation/test sets as shown in Table \ref{tab:split}. The ratio between train and validation/test set is close to $5:1$. Half of the dialogues in validation/test set are grounded in ``unseen'' documents in train set. All experiments were run with 1 V100 GPUs with half precision (FP16) training. More details about experimental settings and hyper-parameters are reported in \ref{app_exp_dpr} and \ref{app_exp_rag}.

\subsection{Baseline Approaches}
Our baseline approaches are based on RAG models \cite{lewis2020retrieval}. For the retriever, we use DPR biencoder pre-trained on Natural Question dataset \footnote{\url{https://github.com/facebookresearch/DPR\#new-march-2021-retrieval-model}}. It contains a question encoder for encoding the dialogue query and a context encoder for encoding document passages. We also fine-tune the DPR bi-encoder using the train and validation set. 
For the generator, we use BART-large pre-trained on CNN dataset. 
To train the retriever and generator end-to-end, we use RAG-Token model, which allows the generator to select content from multiple documents. We found RAG-Token to perform better than RAG-Sequence intuitively and experimentally for our task because MultiDoc2Dial dataset contains many longer agent responses with an average of 22 tokens that might span over multiple passages. We experiment with different retrievers including BM25 and multiple DPR variances.

\subsection{Implementations}

\paragraph{Fine-tuning DPR} To fine-tune DPR, we select positive and negative examples using the train set. For positive examples, we use the grounding annotations, which include the reference document passage information. For negative examples, we use grounding as query to select top one retrieval results as hard negative example; we use dialogue query up to 128 tokens as query, and the top 15 to 25 from retrieval results as the 10 regular negative examples based on best-ranked results by BM25. We use gradient checkpointing to support a large batch size, which we set as 128. This also allows for more in-batch negatives as suggested in \citep{karpukhin-etal-2020-dense,lewis2020retrieval}.

\paragraph{Document Index}
To create document index, we segment documents into passages in two different ways as described in Section \ref{model-doc}. One is to split a document every one hundred tokens, same as the default setting of RAG implementation by Huggingface. We note the token-segmented documents as D\textsuperscript{token}. The other approach is to split a document based on document sections. We note structured-segmented documents as D\textsuperscript{struct}. We use Maximum Inner Product Search (MIPS) to find the top-k documents with Faiss. For indexing method, we use IndexFlatIP``\footnote{\url{https://github.com/facebookresearch/faiss/wiki/Faiss-indexes}} method for indexing.

\paragraph{Dialogue Query Embedding} We combine current turn and history with \texttt{[SEP]} in between as one dialogue query. The query is truncated if longer than maximum source length. To obtain the representation embedding for a dialogue turn, we consider the common \texttt{[CLS]} token embeddings and pooled token embeddings. For the latter, we utilize \texttt{token\_type\_ids} for determining either current turn or history. Then we take average pooling to turn the token embeddings to a fixed-length sequence vector.

\begin{table*}[ht!]
\centering
\begin{tabular}{l c c c  c c c | c c c  c c c}
\hline
& \textbf{F1} & \textbf{EM} & \textbf{BL} & \textbf{@1} & \textbf{@5}&  \textbf{@10}
& \textbf{F1} & \textbf{EM} & \textbf{BL} & \textbf{@1} & \textbf{@5}&  \textbf{@10}
\\ \hline
& \multicolumn{6}{c|}{test set} &  \multicolumn{6}{c}{validation set} \\ 
\hline
D\textsuperscript{token}-bm25 & 28.8 & 2.4 & 12.8 & 20.5 & 41.6 & 50.3 & 28.4 & 2.1 & 13.3 & 19.6 & 41.0 & 48.5 \\
D\textsuperscript{struct}-bm25 & 27.9 & 2.0 & 12.5 & 18.0 & 42.5 & 53.0 & 27.6 & 2.0 & 12.8 & 19.6 & 42.0 & 50.8 \\
\hline
D\textsuperscript{token}-nq & 32.5 & 3.2 & 16.9 & 25.9 & 51.0 & 61.6 & 30.9 & 2.8 & 15.7 & 25.8 & 48.2 & 57.7 \\
D\textsuperscript{struct}-nq & 33.0 & 3.6 & 17.6 & 27.3 & 52.6 & 62.7 & 31.5 & 3.2 & 16.6 & 27.4 & 51.1 & 60.2 \\
\hline
D\textsuperscript{token}-ft & 35.0 & 3.7 & 20.4 & 36.8 & 68.3 & 77.8 & 33.2 & 3.4 & 18.8 & 35.2 & 63.4 & 72.9 \\
D\textsuperscript{struct}-ft & \textbf{36.0} & \textbf{4.1} & \textbf{21.9} & \textbf{39.7} & \textbf{69.3} & \textbf{79.0} & 33.7 & 3.5 & 19.5 & \textbf{37.5} & \textbf{67.0} & \textbf{75.8} \\
D\textsuperscript{token}-rr-cls-ft & 34.9 & 3.5 & 20.0 & 35.7 & 66.8 & 76.6 & 33.2 & 3.4 & 18.4 & 34.7 & 62.2 & 71.8 \\
D\textsuperscript{struct}-rr-cls-ft & \textbf{36.0} & \textbf{4.1} & 21.7 & 38.1 & 67.8 & 77.2 & \textbf{34.1} & \textbf{3.9} & \textbf{19.8} & 36.8 & 65.1 & 74.4 \\
D\textsuperscript{token}-rr-pl-ft & 35.1 & 3.8 & 20.7 & 33.4 & 64.6 & 75.1 & 32.9 & 3.5 & 18.6 & 32.3 & 60.6 & 71.1 \\
D\textsuperscript{struct}-rr-pl-ft & 35.9 & 4.0 & 21.7 & 39.3 & 69.0 & 78.8 & 33.7 & 3.7 & 19.4 & 36.8 & 66.4 & 75.2 \\

\hline
\end{tabular}
\caption{Evaluation results of Task II on agent response generation.}
\label{tab:task2}
\end{table*}

\subsection{Retriever Settings}
We experiment with the following variations for the retriever components.
\begin{itemize}
    \item \textbf{D\textsuperscript{token}} / \textbf{D\textsuperscript{struct}-nq}: uses original pre-trained bi-encoder from DPR. The corresponding document index is based on token/structure-segmented passages.  We consider these setups as baselines.
    \item \textbf{D\textsuperscript{token}-ft} / \textbf{D\textsuperscript{struct}-ft}: uses fine-tuned DPR bi-encoder. The document index is based on token/structure-segmented passages.
    
    \item \textbf{D\textsuperscript{token}-rr-cls-ft} / \textbf{D\textsuperscript{struct}-rr-cls-ft}: uses fine-tuned DPR bi-encoders. We combine the retrieval results of entire dialogue query and only the current turn and select the top-k unique passages. The representation of current turn is based on \texttt{[CLS]} token embeddings. The document index is based on token/structure-segmented passages.
    
    \item \textbf{D\textsuperscript{token}-rr-pl-ft} / \textbf{D\textsuperscript{struct}-rr-pl-ft}: uses fine-tuned DPR bi-encoders. We combine the retrieval results of entire dialogue query and only the current turn and select the top-k unique passages. The representation of current turn is based on pooled token embeddings. The document index is based on token/structure-segmented passages.
\end{itemize}

In addition, we also experiment with BM25 \citep{trotman2014improvements}, noted as\textbf{ D\textsuperscript{token}-bm25}/ \textbf{D\textsuperscript{struct}-bm25)}, where BM25 is used for retrieving top-k passages following the experiment set up in \citet{lewis2020retrieval}.

\subsection{Quantitative Analysis}

\subsubsection{Evaluation Metrics}
We evaluate the passage retrieval results and the generated text for both tasks. For retrieval, we compute recall ($@k$), which measures the fraction of times the correct document is found in the top-$k$ predictions. We evaluate text generation output based on token-level F1 score (F1), Exact Match (EM) \citep{rajpurkar-etal-2016-squad} and SacreBLEU score (BL) \citep{post-2018-call}.

\subsubsection{Passage Retrieval Results}
We first evaluate the performance of BM25, DPR and fine-tuned DPR on a passage retrieval task. The query is the combination of current turn and dialogue history from latest to earliest turn up to 128 tokens. Table \ref{tab:ir} presents the retrieval results on the validation set. BM25 performs better than DPR-nq but worse than DPR-ft. However, it almost shows no difference for D\textsuperscript{token} and D\textsuperscript{struct}. DPR-ft shows significant improvement over DPR-nq. Both DPR-nq and DPR-ft seem to benefit from the document-based structure as they show better performance on D\textsuperscript{struct} than D\textsuperscript{token}.

\subsubsection{Generation Results}
Table \ref{tab:task1} and \ref{tab:task2} present the evaluation results on test and validation set for the two tasks respectively. All numbers on in tables are the \textit{mean} of three runs with different random seeds. We omit the standard deviation numbers as they suggest low variance in our experiments. 
Even though BM25 outperforms DPR-nq in Table \ref{tab:ir}, BM25 performs much worse than DPR-nq for the generation tasks as shown in Table \ref{tab:task1} and \ref{tab:task2}. RAG models with different DPR-based retrievers generally perform better with D\textsuperscript{struct} than D\textsuperscript{token} on the generation tasks. This is consistent with the DPR-based retrieval results in Table \ref{tab:ir}. RAG models with DPR-ft show improvement for both D\textsuperscript{struct} and D\textsuperscript{token} over the ones with DPR-nq, which confirms the importance of positive and negative examples even in small quantity \citep{karpukhin-etal-2020-dense,DBLP:journals/corr/abs-2007-00814}. We also see the retrieval performance gap between D\textsuperscript{token} and D\textsuperscript{struct} is reduced after training the fine-tuned question encoder in RAG. Overall, the retrieval performances for the two tasks seem comparable but generation metric scores for Task II are much lower Task I as the agent responses are free-formed natural language. 

We also experiment with a simple way to re-rank retrieved passages of the entire query based on the retrieved results only based on current turn. We experiment with two types of embeddings for current turn as described earlier. As shown in Table \ref{tab:task1} and \ref{tab:task2}, the re-ranking is not every effective. The difference between reranking with two different kinds of encodings is insignificant. 
In addition, we also evaluate the baselines on unseen domains, where we train the models using the data from three source domains and test on one target unseen domain. For more experiments results on the domain adaptation setup, please see Appendix \ref{appx-val-results}.

\subsubsection{Qualitative Analysis}
To understand the challenges in dialogue grounded in multiple documents and evaluate the data quality, we randomly select some dialog queries from the validation set and examine the queries along with their corresponding retrieved passages by our model. 
We observe certain ambiguities in the dialogue-based and document-based contexts, which we summarize as follows.
\begin{itemize}
    \item {Ambiguity in dialogue queries}: when a user question is under-specific or inquisitive about a higher level topic, it is likely to be quite relevant to multiple document passages. For instance, query \texttt{U8} in \texttt{[Seg-3]} could be linked to different types of benefits described in multiple documents in \texttt{ssa.gov}. 
    \item {Ambiguity in document content}: when certain passages with very similar topics in different documents, they could be duplicate or different in context. For instance, same question could be addressed in an FAQ page and in another article regarding a different specific criterion.
\end{itemize}

\paragraph{Human evaluations on ambiguity in questions} We ask human experts to identify whether a dialogue inquiry turn is ambiguous based on its dialogue history and the associated documents. We consider a dialogue query as `ambiguous' if it is likely to be relevant to a broad range of domain knowledge; otherwise as `unambiguous'. Firstly, we ask the the annotators to annotate the queries based on their understanding of dialogue context and the domain knowledge. Secondly, we reveal the reference document passage and another most relevant document passage retrieved by our models and ask them to annotate the queries again. We randomly select 100 dialogue inquiry turns and assign them to two experts with 20\% overlap. The Cohen's kappa agreement score is 0.85. In the first setting, 15\% of the turns are labeled as `ambiguous'. In the second setting, after revealing the relevant passages, 20\% turns are considered `ambiguous'. 

Such ambiguities are generally inherent to open retrieval settings \citep{zhu2021retrieving,min-etal-2020-ambigqa}. In practice, it would require the conversational agents to ask follow-up questions for clarification based on dialogue history and retrieved passages for providing a more fair and informative answer. The current version of MultiDoc2Dial dataset can be further enhanced by adding agent turns for asking clarification questions based on different levels such as highest title (topic) level, sub-title (sub-topic) level or even finer level, which we leave for near future work.

\section{Related Work}
\subsection{Document-grounded Dialogue and Conversational Question Answering} 

Our work is closely related to the recent work on document-grounded dialogue and conversational machine reading comprehension tasks, such as Doc2Dial \citep{feng-etal-2020-doc2dial}, ShARC \citep{saeidi-etal-2018-interpretation} and DoQA \citep{campos-etal-2020-doqa}. The primary goal of these papers is to provide an answer, or a dialogue response based on a single given document or text snippet. In contrast to the closed-book setting of these works, our task is in an open-book setting, which aims to address more realistic scenarios in goal-oriented dialogues where the associated document content are unknown and likely more than one documents.

Our work is built on Doc2Dial, which is a goal-oriented dialogue modeling task based on a single document. Our dataset shares the same set documents and annotation scheme as Doc2Dial. However, we aim to further the challenge by dealing with cases when the document-level topic shifts through a dialogue. Thus, we proposed a new data construction approach, data tasks, and baseline approaches in an end-to-end setting.

\subsection{Open Domain Question Answering}

Our proposed data and tasks are also related to open domain question answering \citep{chen-etal-2017-reading,lee-etal-2019-latent,kwiatkowski-etal-2019-natural,min-etal-2020-ambigqa,qu2020open,mao2020generationaugmented,zhu2021retrieving, izacard2021distilling, li2021graph, xiong2021answering, yu2021fewshot}. In particular, our work is closely related to the recently proposed open-retrieval conversational question answering (OR-CQA) setting for QuAC dataset, i.e., OR-QuAC \citep{qu2020open}. 
To the best of our knowledge, the search queries for tasks are mostly created based on Wikipedia articles. In OR-QuAC \citep{qu2020open}, all turns in one conversation are grounded in passages from the Wikipedia page of the given entity, i.e., there are no multiple documents involved in a dialogue. 
In contrast, our task proposes to model dialogues grounded in multiple documents. The documents are of diverse writing styles from four real user-facing websites. In addition to the difference in document data, MultiDoc2Dial provides more types of dialogue query based on a richer set of dialogue acts. Table \ref{tab:comparison} provides a comparison of several most related datasets and tasks in different aspects including whether the setup is open-book or not, the dialogues are goal-oriented or not, the grounding is annotated or not, the associated text is full document or not and each dialogue corresponds to multiple documents or not. Our work is the only one that covers all the characteristics.

\subsection{Discourse Segmentation}
This work is also largely related to discourse segmentation tasks \citep{arguello-rose-2006-topic,zhang2019topic,mele2020topic,gao-etal-2020-discern,xing-carenini-2021-improving}, which aims to identify the change of topic in a dialogue. This is a very important task towards modeling of goal-oriented dialogues in general. Some papers, such as \citet{arguello-rose-2006-topic,hsueh-etal-2006-automatic,xing-carenini-2021-improving}, focus on the task of modeling and predicting segmentation; some papers such as \citet{gao-etal-2020-discern} use explicit segmentation as the input of downstream dialogue modeling tasks on a machine reading comprehension dataset, ShaRC. Our task is closely related to the latter, albeit in an open book setting and with an end-to-end modeling approach that encodes the dialogue segmentation information implicitly. We leave the more explicit dialogue segmentation modeling for future work. 

\section{Conclusion and Future Work}
We introduced MultiDoc2Dial, a new task and dataset that deals with goal-oriented dialogues that have multiple sub-goals corresponding to different documents. 
We proposed two tasks for predicting next agent turn that we formulate as generation tasks. We presented strong baseline approaches based on retriever-reader architecture and experimented with different variances of neural retrievers. For future work, we aim to address the ambiguity in the open-book dialogue modeling.

\section{Ethical Consideration}
One primary motivation of the paper is to provide data instances that simulate how real human users converse with agents to seek information. Such data is essential for training neural models to build conversational systems that could assist various end users to access information in real-life domains such as social benefit websites. However, such a dataset is largely unavailable for research and development. Since we create the dataset via crowdsourcing, one potential ethical concern is that it could be potentially biased or distant from the real user queries. To address such concerns, we try to identify qualified contributors with different backgrounds and train them via several rounds of tasks. In addition, we provide various examples in the instruction and feedback to contributors during the crowdsourcing task.

\bibliography{anthology,custom}
\bibliographystyle{acl_natbib}

\clearpage
\newpage
\appendix

\section{Data Construction}
\label{appx_crowd}

\begin{figure*}
     \centering
     \begin{subfigure}[b]{0.9\textwidth}
         \centering
         \includegraphics[width=\textwidth]{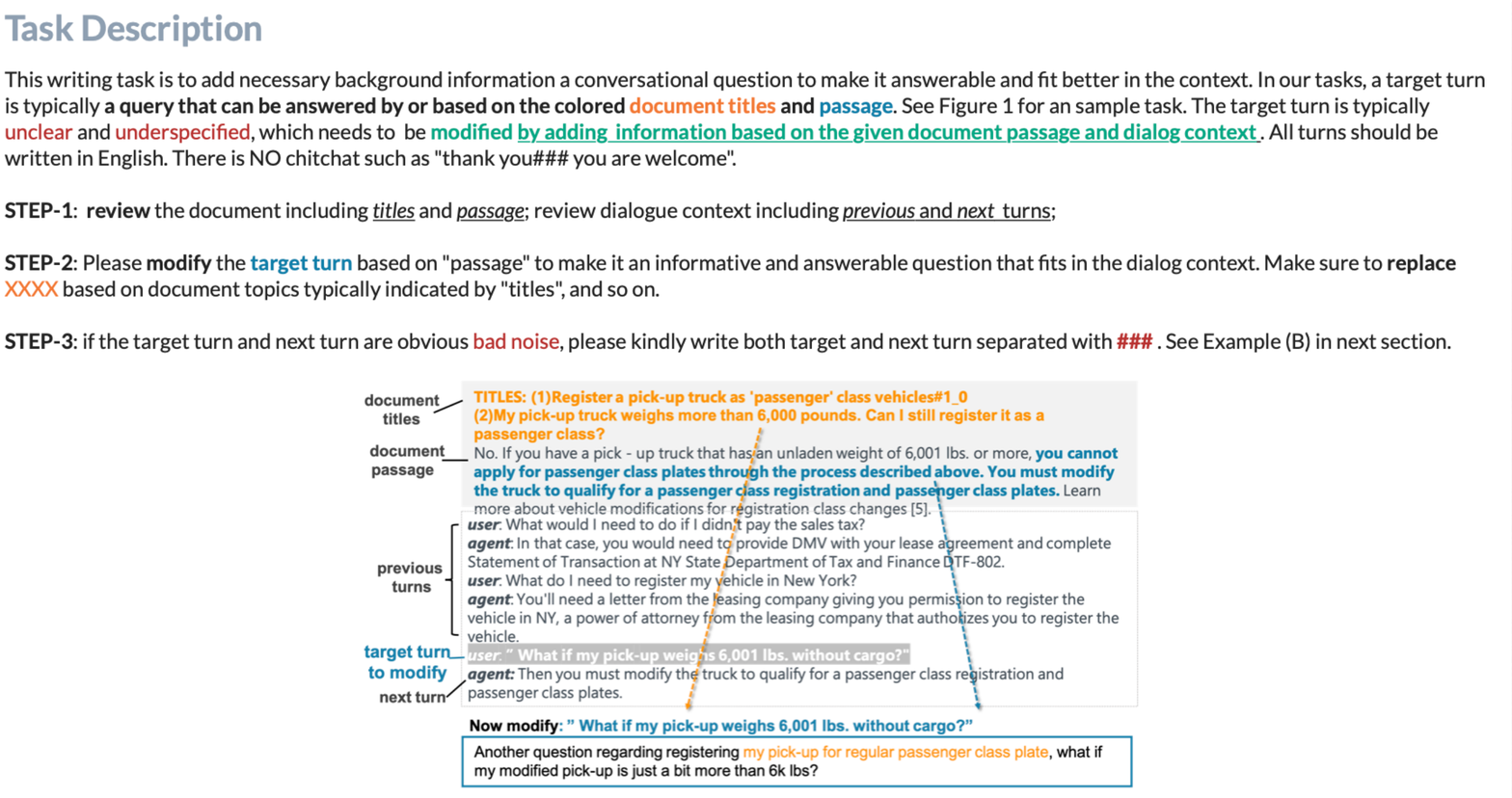}
         \caption{The screenshot of task description and data collection interface of the crowdsourcing task.}
         \label{fig:crowd1}
     \end{subfigure}
     \hfill
     \begin{subfigure}[b]{0.9\textwidth}
         \centering
         \includegraphics[width=\textwidth]{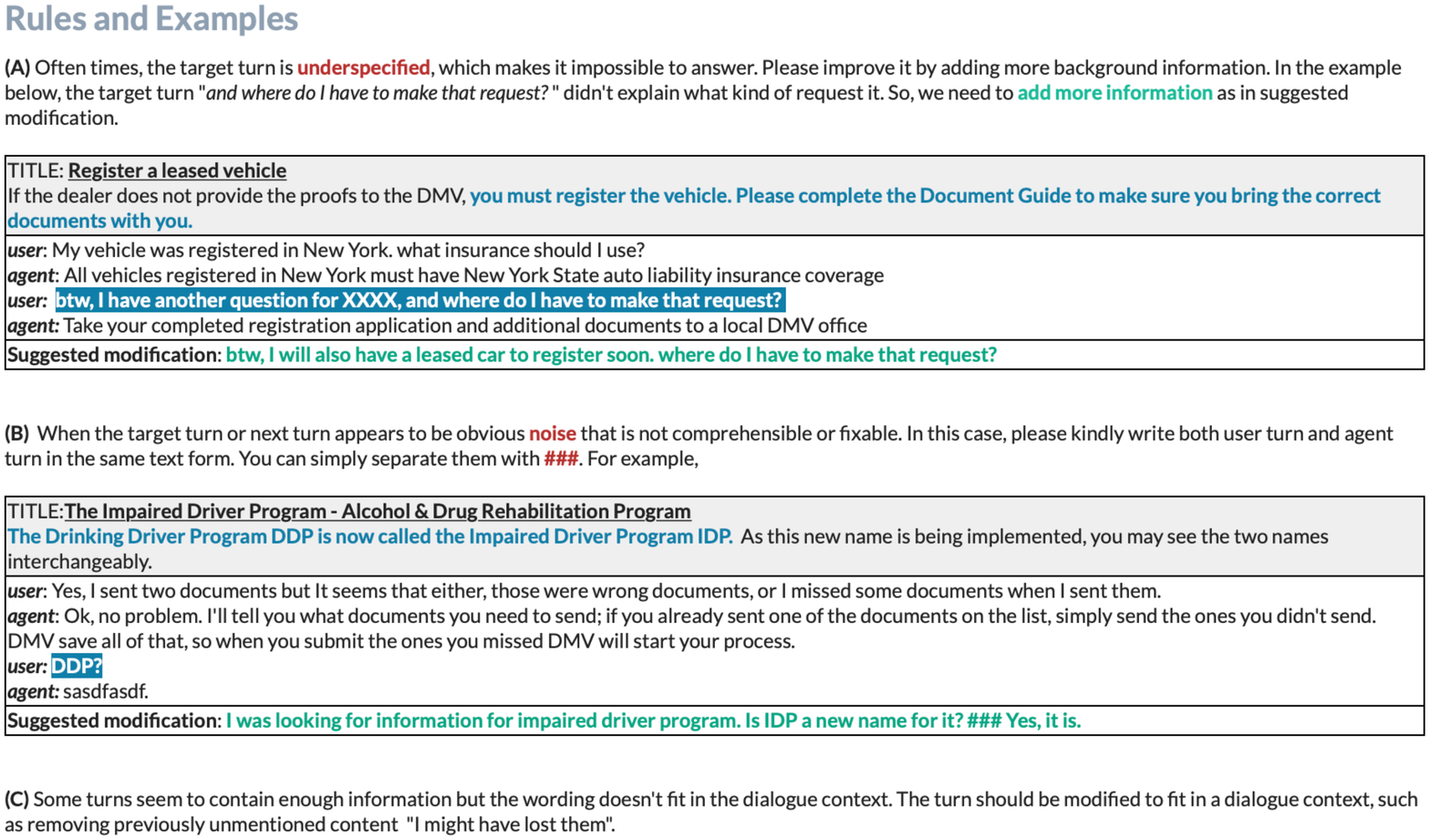}
         \caption{The screenshot of part of the rules and examples of the crowdsourcing task.}
         \label{fig:crowd2}
     \end{subfigure}
\end{figure*}

For crowdsourcing, we filter out less qualified contributors by adding template-based placeholders in the writing task for detecting bad performances, which seems effective. Most contributors seem to be able to improve either user query or agent response, sometimes both together based on document context. However, we do find the writings from original Doc2Dial dataset appear a bit more natural with more personalized information. We suspect that it could be easier to add such information in the writing if the conversation is built from scratch based on a single document while the contributors write the dialogue history themselves. We provide feedback accordingly to the crowd to address the issue. We observe that the crowd contributors hardly reject any task. For quality control, we also manually review and re-collect data during the data collection process. For the instruction, interface, rules and examples for data collection via crowdsourcing, please see Figure \ref{fig:crowd1} and \ref{fig:crowd2} for reference.

\section{Experiments}
\label{app_exp}
The implementation is in PyTorch. For fine-tuning RAG \citep{lewis2020retrieval}, we follow the example\footnote{\url{https://github.com/huggingface/transformers/blob/master/examples/research_projects/rag/finetune_rag.py}} from HuggingFace and fine-tune it on our dataset for 16 epochs. For fine-tuning DPR \citep{karpukhin-etal-2020-dense}, we train DPR-nq on our dataset using facebookresearch/DPR\footnote{\url{https://github.com/facebookresearch/DPR}}. Then, we integrate fine-tuned bi-encoder in RAG model \texttt{facebook/rag-token-nq}\footnote{\url{https://huggingface.co/facebook/rag-token-nq}}. For pre-trained DPR, we use the bi-encoder model trained on NQ dataset only from Facebook DPR checkpoint\footnote{\url{https://github.com/facebookresearch/DPR\#new-march-2021-retrieval-model}}. We train the models for 10 epochs and evaluate using the last checkpoint.

\subsection{Hyperparameters for fine-tuning DPR}
\label{app_exp_dpr}
We fine-tune DPR for 50 epochs with a batch size of 128. We use a learning rate of 2e-05 using Adam, linear scheduling with warmup and dropout rate of 0.1. We set the max encoder sequence length to 128 consistent with the RAG model. We also use one additional BM25 negative passage per question in addition to in-batch negatives. We use gradient checkpointing to support a large batch size as 128. The performance of checkpoint at epoch 50 and 80 are very close. We use checkpoint at 50 epoch across experiments.

\subsection{Hyperparameters for fine-tuning RAG}
\label{app_exp_rag}
Hyper-parameters used for fine-tuning as as follows.
\begin{verbatim}
train_batch_size=8
eval_batch_size=2
max_combined_length=300
max_source_length=128
max_target_length=50
val_max_target_length=50
test_max_target_length=50
label_smoothing=0.1
dropout=0.1
attention_dropout =0.1
weight_decay=0.001
adam_epsilon=1e-08
max_grad_norm=0.1
lr_scheduler=polynomial
learning_rate=3e-05
warmup_steps=500
gradient_accumulation_steps=1 
\end{verbatim}

\begin{table}
\begin{center}
\begin{tabular}
{ m{1.5cm} m{2cm} m{2cm}}
\hline
\textbf{} & \textbf{Retriever} & \textbf{Reader}  \\
\hline
\multirow{2}{1.5cm}{DPR } & BERT-base & - \\
  & 2X100M &  - \\
  \hline
\multirow{2}{1.5cm}{RAG }  & BERT-base & BART-large\\
& 2X100M & 330M \\
\hline
\end{tabular}
\caption{The number of parameters of models.}
\label{tab:num_para}
\end{center}
\end{table}

\subsection{Experiment Results}
\label{appx-val-results}
\paragraph{Fine-tuned DPR} For fine-tuning DPR, we experiment with different ways of obtaining hard negative examples. One is using the grounding of a dialogue query (grounding) as query, the other is using combined dialogue utterances (query) as query. The results turn out comparable as shown in Table \ref{tab:ir-epochs}.

\paragraph{Domain Adaptation Setup} We also experiment with domain adaption setup, where the train and validate splits are based on all data from the three domains and the test split is based on one unseen domain. The domain information is considered when retrieving relevant passage.  The ratio of the number of examples in train and validation is $5:1$. Table \ref{tab:ir-domain} presents the retrieval results of different source and target unseen domains, which is comparable to Table \ref{tab:ir}. However, the EM and F1 scores in Table \ref{tab:task-test-domain} are much lower comparing to the setup without any unseen domain in Table \ref{tab:task1} and \ref{tab:task2}, which confirms that the domain adaptation setup is indeed challenging.

\begin{table}
\begin{center}
\begin{tabular}
{ c c c c c}
\hline
\textbf{Unseen} & \textbf{Doc-Seg} & \textbf{R@1} &  \textbf{R@5} & \textbf{R@10} \\
\hline
\multirow{2}{3em}{ssa} & token & 50.0 & 69.6 & 73.8\\
  & struct & 65.7 & 86.3 & 90.7 \\
\hline
\multirow{2}{3em}{va} & token & 48.0 & 68.0 & 73.8\\
  & struct & 64.0 & 85.2 & 90.3 \\
\hline
\multirow{2}{3em}{dmv} & token & 51.3 & 69.7 & 75.1\\
  & struct & 67.6 & 86.2 & 91.2 \\
\hline
\multirow{2}{3em}{studentaid} & token & 49.7 & 69.4 & 75.0\\
  & struct & 66.0 & 85.8 & 91.2 \\
\hline

\end{tabular}
\caption{Retrieval results on validation set for the domain adaption setup.}
\label{tab:ir-domain}
\end{center}
\end{table}

\begin{table}
\centering
\begin{center}
\begin{tabular}
{c c c c c c}
\hline
\textbf{Neg.} & \textbf{Doc-Seg} & \textbf{@1} &  \textbf{@5} & \textbf{@10} & \textbf{@50} \\
\hline
& \multicolumn{5}{c}{ 40 epochs} \\
\hline
\multirow{2}{3em}{query} & token & 36.5 & 58.2 & 65.0 & 77.3\\
{} & struct  & 48.6 & 71.1  & 78.9 & 91.7 \\

\multirow{2}{3em}{grounding}  &  token & 37.0 & 58.8 & 65.7 & 76.5\\
{} & struct  & 47.7 & 70.2 &  77.7 & 91.4\\
\hline
& \multicolumn{5}{c}{ 50 epochs} \\
\hline
\multirow{2}{3em}{query} & token & 36.5 & 57.4 & 64.6 & 77.1\\
{} & struct  & 49.0 & 72.3  & 80.0 & 92.5 \\

\multirow{2}{3em}{grounding}  &  token & 35.9 & 57.6 & 65.6 & 76.8\\
{} & struct  & 47.8 & 71.0 &  78.8 & 92.0\\
\hline
\end{tabular}
\caption{Retrieval results on validation set with respect to different DPR settings.}
\label{tab:ir-epochs}
\end{center}
\end{table}

\begin{table*}
\centering
\begin{tabular}{l l c c c  c c c}
\multicolumn{8}{c}{Task I: grounding generation} \\
 \hline
 & & \textbf{F1} & \textbf{EM} & \textbf{BLEU}
 & \textbf{R@1} &  \textbf{R@5}&  \textbf{R@10}
 \\
\hline
 \multirow{4}{4em}{ssa} & D\textsuperscript{token}-ft & 29.0  &  8.9  &  20.8  &  15.3  &  39.8  &  52.5\\
 & D\textsuperscript{struct}-ft & 27.2  &  7.7  &  17.6  &  15.9  &  45.9  &  61.3\\
 & D\textsuperscript{token}-rr-pl-ft & 29.0  &  9.0  &  20.7  &  15.3  &  39.9  &  52.5\\
 & D\textsuperscript{struct}-rr-pl-ft & 27.3  &  7.7  &  18.0  &  15.9  &  46.0  &  61.4\\
 \hline
 \multirow{4}{4em}{va} & D\textsuperscript{token}-ft & 31.4  &  10.9  &  23.0  &  32.1  &  57.2  &  66.7\\
 & D\textsuperscript{struct}-ft & 31.2  &  11.8  &  23.5  &  34.1  &  61.1  &  70.5\\
 & D\textsuperscript{token}-rr-pl-ft & 31.2  &  11.0  &  23.1  &  32.1  &  57.2  &  66.7\\
 & D\textsuperscript{struct}-rr-pl-ft & 31.0  &  11.7  &  23.5  &  34.1  &  61.1  &  70.5\\
 \hline
 \multirow{4}{4em}{dmv} & D\textsuperscript{token}-ft & 29.4  &  9.6  &  19.6  &  31.7  &  58.3  &  68.8\\
 & D\textsuperscript{struct}-ft & 28.6  &  9.8  &  20.7  &  34.9  &  61.5  &  72.2\\
 & D\textsuperscript{token}-rr-pl-ft & 29.3  &  9.4  &  19.5  &  31.7  &  58.3  &  68.8\\
 & D\textsuperscript{struct}-rr-pl-ft & 28.7  &  9.9  &  20.8  &  34.9  &  61.5  &  72.2\\
 \hline
 \multirow{4}{4em}{studentaid} & D\textsuperscript{token}-ft & 29.6  &  9.6  &  21.2  &  34.6  &  62.1  &  71.8\\
 & D\textsuperscript{struct}-ft & 28.9  &  9.6  &  20.8  &  38.3  &  66.6  &  76.2\\
 & D\textsuperscript{token}-rr-pl-ft & 29.3  &  9.4  &  21.2  &  34.6  &  62.1  &  71.8\\
 & D\textsuperscript{struct}-rr-pl-ft & 29.0  &  9.6  &  20.8  &  38.3  &  66.6  &  76.2\\
\hline
\end{tabular}
\caption{Evaluation results of Task I on test set of unseen domains.}
\label{tab:task-test-domain}
\end{table*}

\begin{table*}
\centering
\begin{tabular}{l l c c c  c c c}
\multicolumn{8}{c}{Task II: response generation} \\
 \hline
 & & \textbf{F1} & \textbf{EM} & \textbf{BLEU}
 & \textbf{R@1} &  \textbf{R@5}&  \textbf{R@10}
 \\
\hline
 \multirow{4}{4em}{ssa} & D\textsuperscript{token}-ft & 28.1  &  1.1  &  13.3  &  15.3  &  39.8  &  52.5\\
 & D\textsuperscript{struct}-ft & 28.0  &  1.1  &  13.2  &  15.9  &  45.9  &  61.3\\
 & D\textsuperscript{token}-rr-pl-ft & 28.2  &  1.1  &  13.4  &  15.3  &  39.9  &  52.5\\
 & D\textsuperscript{struct}-rr-pl-ft & 28.0  &  1.1  &  13.3  &  15.9  &  46.0  &  61.4\\
 \hline
  \multirow{4}{4em}{va} & D\textsuperscript{token}-ft & 30.4  &  2.4  &  17.2  &  32.1  &  57.2  &  66.7\\
 & D\textsuperscript{struct}-ft & 31.2  &  2.4  &  18.6  &  34.1  &  61.1  &  70.5\\
 & D\textsuperscript{token}-rr-pl-ft & 30.7  &  2.5  &  17.4  &  32.1  &  57.2  &  66.7\\
 & D\textsuperscript{struct}-rr-pl-ft & 31.2  &  2.3  &  18.6  &  34.1  &  61.1  &  70.5\\
 \hline
  \multirow{4}{4em}{dmv} & D\textsuperscript{token}-ft & 28.9  &  1.4  &  13.8  &  31.7  &  58.3  &  68.8\\
 & D\textsuperscript{struct}-ft & 29.5  &  1.7  &  14.9  &  34.9  &  61.5  &  72.2\\
 & D\textsuperscript{token}-rr-pl-ft & 28.8  &  1.3  &  13.8  &  31.7  &  58.3  &  68.8\\
 & D\textsuperscript{struct}-rr-pl-ft & 29.6  &  1.6  &  15.0  &  34.9  &  61.5  &  72.2\\
 \hline
  \multirow{4}{4em}{studentaid} & D\textsuperscript{token}-ft & 29.0  &  0.9  &  13.6  &  34.6  &  62.1  &  71.8\\
 & D\textsuperscript{struct}-ft & 28.8  &  1.0  &  14.0  &  38.3  &  66.6  &  76.2\\
 & D\textsuperscript{token}-rr-pl-ft & 29.1  &  1.0  &  13.8  &  34.6  &  62.1  &  71.8\\
 & D\textsuperscript{struct}-rr-pl-ft & 28.8  &  0.9  &  13.9  &  38.3  &  66.6  &  76.2\\

\hline
\end{tabular}
\caption{Evaluation results of Task II on test set of unseen domains.}
\label{tab:task-val-domain}
\end{table*}

\end{document}